\documentclass[letterpaper, 10 pt, conference]{ieeeconf}  % Comment this line out if you need a4paper

\IEEEoverridecommandlockouts                              % This command is only needed if 
\usepackage{booktabs}
\usepackage{multirow}
\usepackage{subfigure} % 子图包
\usepackage{booktabs}
\usepackage{graphicx}
\usepackage[namelimits]{amsmath} %数学公式
\usepackage{amssymb}             %数学公式
\usepackage{amsfonts}            %数学字体
\usepackage{mathrsfs}            %数学花体

\title{\LARGE \bf
Multi-modal Motion Prediction with Transformer-based Neural Network for Autonomous Driving
}

\author{Zhiyu Huang, Xiaoyu Mo, and Chen Lv$^{*}$,~\IEEEmembership{Senior Member, IEEE}% <-this % stops a space
\thanks{Z. Huang, X. Mo, and C. Lv are with the School of Mechanical and Aerospace Engineering, Nanyang Technological University, 639798, Singapore. (E-mails: {\tt\small zhiyu001@e.ntu.edu.sg}, {\tt\small xiaoyu006@e.ntu.edu.sg}, {\tt\small lyuchen@ntu.edu.sg}.)}%
\thanks{This work was supported in part by A*STAR Grant (No. 1922500046), A*STAR AME Young Individual Research Grant (No. A2084c0156), and SUG-NAP Grant (No. M4082268.050) of Nanyang Technological University, Singapore.}% <-this % stops a space
\thanks{$^{*}$Corresponding author: C. Lv}
}

\begin{document}
\maketitle
\thispagestyle{empty}
\pagestyle{empty}

%%%%%%%%%%%%%%%%%%%%%%%%%%%%%%%%%%%%%%%%%%%%%%%%%%%%%%%%%%%%%%%%%%%%%%%%%%%%%%%%
\begin{abstract}

Predicting the behaviors of other agents on the road is critical for autonomous driving to ensure safety and efficiency. However, the challenging part is how to represent the social interactions between agents and output different possible trajectories with interpretability. In this paper, we introduce a neural prediction framework based on the Transformer structure to model the relationship among the interacting agents and extract the attention of the target agent on the map waypoints. Specifically, we organize the interacting agents into a graph and utilize the multi-head attention Transformer encoder to extract the relations between them. To address the multi-modality of motion prediction, we propose a multi-modal attention Transformer encoder, which modifies the multi-head attention mechanism to multi-modal attention, and each predicted trajectory is conditioned on an independent attention mode. The proposed model is validated on the Argoverse motion forecasting dataset and shows state-of-the-art prediction accuracy while maintaining a small model size and a simple training process. We also demonstrate that the multi-modal attention module can automatically identify different modes of the target agent's attention on the map, which improves the interpretability of the model.

\end{abstract}

%%%%%%%%%%%%%%%%%%%%%%%%%%%%%%%%%%%%%%%%%%%%%%%%%%%%%%%%%%%%%%%%%%%%%%%%%%%%%%%%
\section{INTRODUCTION}

Accurately predicting traffic participants’ future trajectories is critical for autonomous vehicles to make safe, informed, and human-like decisions \cite{huang2021driving}, especially in complex traffic scenarios. However, motion prediction is a remarkably challenging task due to the complicated dependencies of agents' behaviors on the road structure and interactions among agents in addition to their kinematics, as well as the inherent uncertainty and multi-modality of their intentions.

One major challenge is how to represent the driving environment in the prediction model, including encoding road structure and agent interactions, in addition to the target's physical states. The difficulty of representing interactions or map data in explicit rules or parametric models gives rise to deep neural network-based methods \cite{mozaffari2020deep}, which are able to handle high-dimensional map data and represent the agent interaction patterns by learning from human driving data. In this paper, we employ a graph-based structure to represent the relationship between the interacting agents, where all the agents are treated as nodes and each surrounding agent is connected to the target agent, and a Transformer-based encoder to model the relationship among the target agent and its surrounding agents. The multi-head attention in the Transformer layer can help extract different aspects of the agent interactions. As for extracting the relationship between the map and the target agent, we utilize the vectorized map representation \cite{gao2020vectornet} and a lane set-based map structure consisting of a list of waypoints from different lanes, which provides a higher map resolution. To model the target agent's different aspects of attention on the different lane segments, we propose the multi-modal Transformer encoder that can extract the different modes of agent-map relations. The rationals and details are given below.
% assuming a fully connected graph where all the waypoints are connected to the target agent node

The motion prediction model should be capable of outputting multiple possible trajectories, and we can notice that the uncertainty of agent behaviors predominantly comes from their different targets on the road \cite{zhao2020tnt}. Therefore, in modeling the attention between the target agent and the map, we propose to modify the multi-head attention mechanism in the Transformer to multi-modal attention. The multi-head attention is designed to extract multiple possible relationships and richer interpretations between the inputs, and all of the attention heads are then merged to produce a final output. Instead of combining them to output a single result, we propose to directly output the result of each independent attention head, and each separate agent-map attention result is used for predicting a possible trajectory. The intuition behind this is that each attention head can represent a different relationship between the target agent and different segments of the map, which affects their future trajectories. Moreover, to ensure the diversity of the multiple attention heads, only the head that outputs the closest trajectory to the ground-truth one gets updated during training. 

The proposed motion prediction model is succinct, which encompasses only two cross-attention Transformer layers for modeling agent interactions and map attention respectively, other than the trajectory and map encoders and prediction head, enabling it to be easy to train and deploy while maintaining satisfactory prediction accuracy. The main contributions of this paper are listed as follows.

\begin{enumerate}
\item We propose a Transformer-based network using the waypoint-based map structure for multi-modal trajectory prediction. The proposed multi-modal attention Transformer layer is shown to be able to capture the different modes of agent-map relations, thus bringing better accuracy and interpretability.

\item We validate the proposed method on a large-scale real-world driving dataset and the results reveal that the proposed method with a simple structure and training process can achieve competitive accuracy compared to other state-of-the-art methods.

\item We investigate the performance of using lane-based and waypoint-based map structure with the proposed prediction network, as well as the improvement of the proposed multi-modal attention mechanism.
\end{enumerate}

\section{RELATED WORK}

\subsection{Encoding Map and Interaction}
The most common method of encoding map and agent interaction information is to rasterize the driving scene into bird-eye-view images, which contain the information of relationships among agents and road structure. Such environment representation can be effectively processed by convolutional neural networks (CNNs), which have been used in many motion prediction-related works \cite{cui2019multimodal, salzmann2020trajectron++, dong2021multi}. However, the drawback of using rasterized images and CNNs is that the image structure is overly complex for driving environment representation and thus it requires a larger network to process and more computation and data to train the network. More recently, a notably compact vector map representation \cite{gao2020vectornet, liang2020learning} has been proposed, which can significantly reduce the computation burden. VectorNet \cite{gao2020vectornet} treats the lanes on the map and agent historical trajectories as a set of polylines, and models them as a global fully-connected interaction graph, which is proposed by a graph neural network (GNN) to encode the map and agent information. However, putting all information including lanes and agents in a single graph makes training hard and inefficient, and thus we utilize two separate Transformer-based layers to encode agent interaction and agent's attention on map, respectively. On the other hand, LaneGCN \cite{liang2020learning} proposes to organize the lanes on the map into a lane graph considering the spatial connectivity, and then use the graph convolutional network (GCN) to encode the topology of the map. However, using lane-level features may decrease the map resolution, and thereby we propose to use lane set-based map representation, which is a set of waypoints from different lanes with minimal information loss.

\subsection{Multi-modal Prediction}
To realize multi-modal prediction, i.e., predicting multiple possible future trajectories, some generative modeling approaches, such as conditional variational autoencoder (CVAE) \cite{lee2017desire, ivanovic2020multimodal} and generative adversarial network (GAN) \cite{gupta2018social}, are employed. However, such generative methods are hard to train and infer because the model needs to be sampled many times to recover a plausible distribution over future behaviors. Some other works propose to predict a set of trajectories \cite{cui2019multimodal, liang2020learning, ye2021tpcn} using the variety loss \cite{thiede2019analyzing}. However, they use the same extracted feature vector to output multiple trajectories, which is not intuitive and lacks interpretability on the outputs. Anchor-based methods \cite{chai2020multipath, phan2020covernet, song2021learning} can provide better interpretability, feasibility, and diversity on the results, but their predictions are restricted to a predefined set, obtained by clustering from the data or generated by a model, which may impede the prediction accuracy. On the other hand, goal-based \cite{zhao2020tnt} or proposal-based methods \cite{zhang2020map, liu2021multimodal, fang2020tpnet} have been widely used due to their superior accuracy and interpretability. The trajectory predictions are conditioned on the possible long-term goals or proposals on the map, which brings diversity and also flexibility to the model outputs. Nevertheless, the goals or proposals are still manually selected, which is laborious in data processing and needs careful design. Different from these methods above, in this work, we propose to use the multi-modal Transformer layer to learn to attend to different segments of the map and produce diverse possible trajectories accordingly in an end-to-end manner, which can simplify the training process and maintain accuracy, interpretability, and flexibility.

\section{Multi-modal Motion Prediction Framework}
\subsection{Problem Formulation} 
The task of motion prediction is to predict the possible future trajectories of a target agent over a time horizon $T_f$ based on its historical states over a time period $T_h$ and environmental context information. The input $\mathbf{X}$ to the prediction model consists of the historical dynamic states of the target agent ($S_{0}$) and its surrounding agents ($S_{1}, \dots, S_{N}$), as well as the current environment information $\mathcal{M}$. Without loss of generality, we assume that there are $N$ surrounding agents (e.g., vehicles, pedestrians, and cyclists) around the target agent, however, the number of surrounding agents can be varied in different situations. The output of the prediction model $\Hat{\mathbf{Y}}$ is $K$ trajectories, each consisting of a discrete sequence of 2D coordinates $\xi_j$, denoting the future positions of the target agent, as well as the corresponding probability $p_j$. Mathematically, the problem is formulated as:
\begin{equation}
\begin{aligned}
\Hat{\mathbf{Y}} &= f \left( \mathbf{X} | \theta \right), \\
\mathbf{X} &= \left\{ S_{0}, S_{1}, \dots, S_{N}, \mathcal{M} \right\}, \\
\Hat{\mathbf{Y}} &= \left\{ \xi_j , p_j \right\}_{j=1}^{K}, \\
\xi_j &= \left\{ (x^t_{j}, y^t_{j}) | t \in \left\{ t_{0}+1, \dots, t_{0}+T_{f} \right\} \right\},
\end{aligned}
\end{equation}
where $\theta$ denotes the parameters of the prediction model $f$, $S_{i}= \left\{ s_{i}^{t} | t \in \left\{ {t_0-T_h}, \dots, {t_0} \right\} \right\}$ and $s_{i}^{t}$ is the dynamic state of the agent $i$ at timestep $t$, $(x^t_{j}, y^t_{j})$ is the coordinate of the $j$th predicted trajectory at timestep $t$, and $t_0$ is the current time step.

\begin{figure*}[htp]
    \centering
    \includegraphics[width=\linewidth]{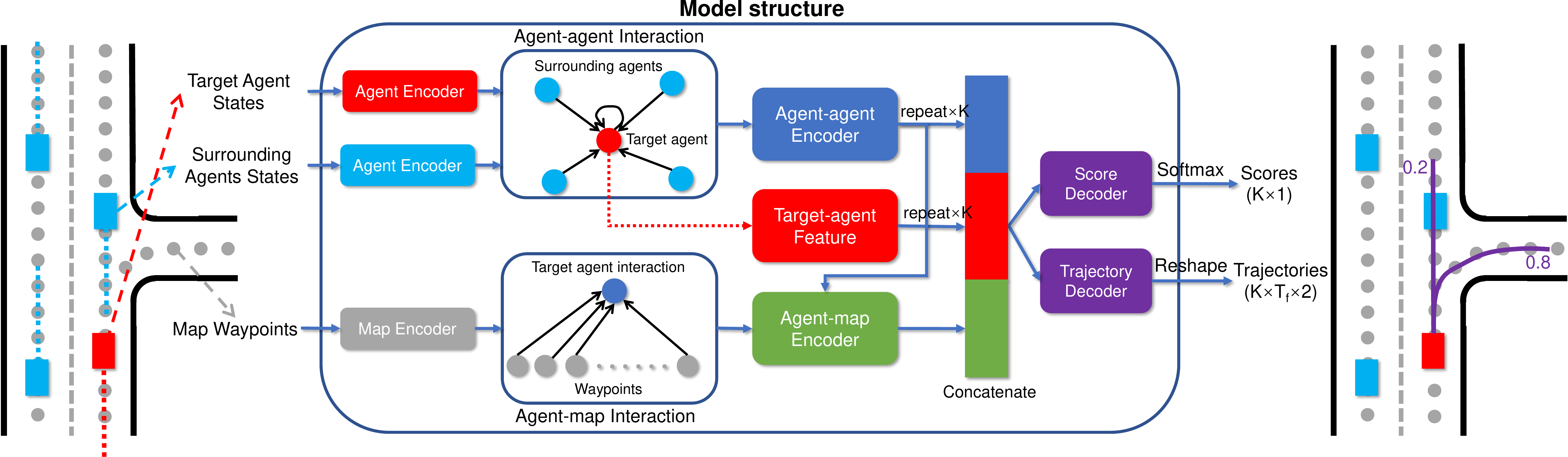}
    \caption{An overview of our proposed motion prediction model. The agent encoder and map encoder are used to extract the features of agents and map waypoints, respectively. The agent-agent encoder is employed to model the relationship among interacting agents, and the map-agent encoder to model the relationship between the target agent with interaction feature and the waypoints on the map. Finally, the interaction feature, target's agent historical feature, and agent-map attention feature are concatenated and passed through the trajectory and score decoders, to output the predicted trajectories and their scores.}
    \label{fig:fig.1}
\end{figure*}

\begin{figure}[htp]
    \centering
    \includegraphics[width=\linewidth]{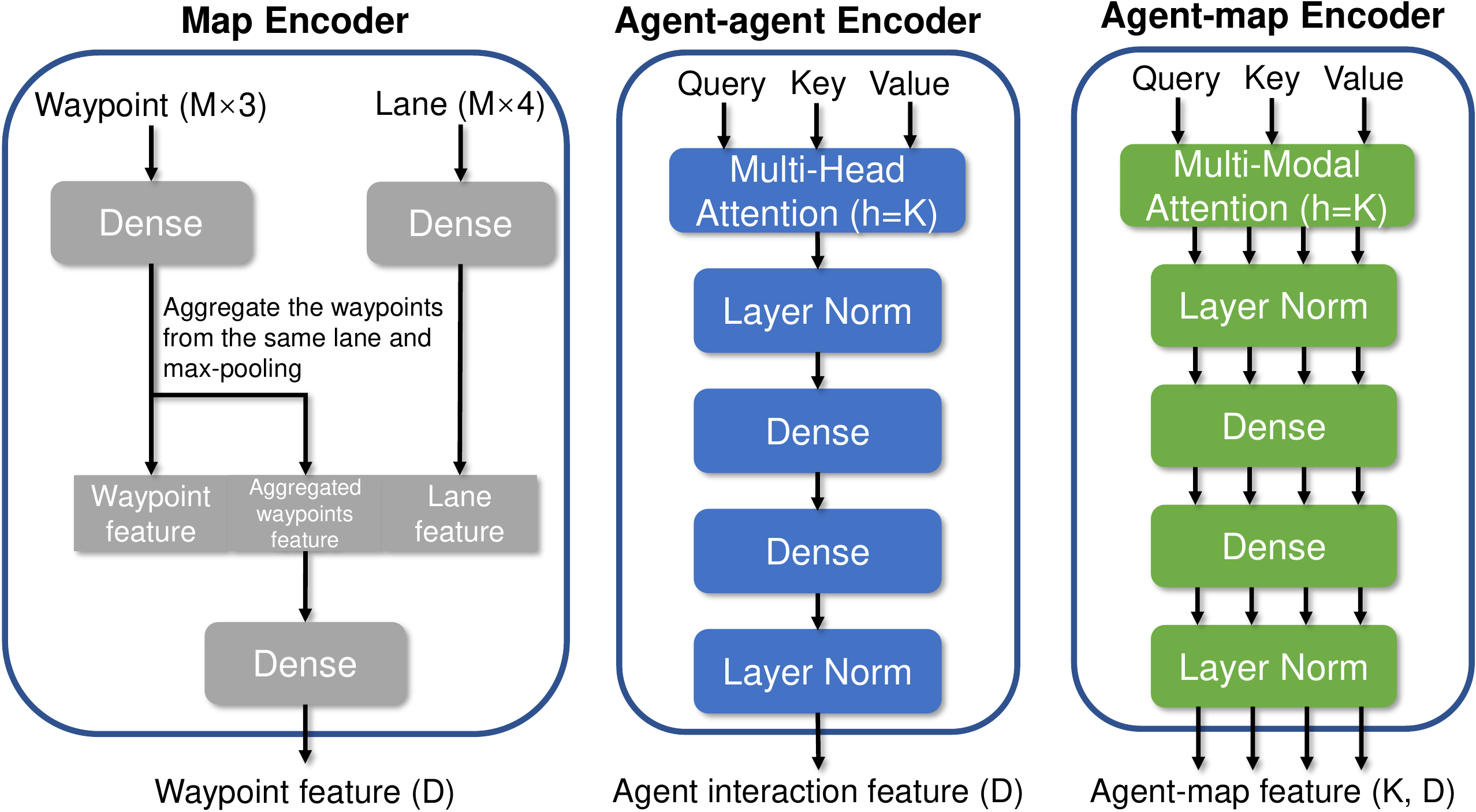}
    \caption{The detailed structures of the map encoder, agent-agent encoder, and agent-map encoder in the prediction model.}
    \label{fig:fig.2}
\end{figure}

\subsection{Prediction Framework}
The proposed motion prediction framework is illustrated in Fig. \ref{fig:fig.1}, which encompasses four main parts. First, the map and agent encoders translate the low-dimensional raw state inputs to high-dimensional feature vectors. Then, the agent-agent encoder is used to capture the relationship between the interacting agents, and the agent-map encoder models the target agent's attention on different segments of the map. Finally, the interaction feature, map attention feature, and dynamic feature of the target agent are concatenated and passed through the scored trajectory decoder to generate possible trajectories with associated probabilities. The detailed structures of the key components, i.e, the map encoder, agent-agent encoder, and agent-map encoder are illustrated in Fig. \ref{fig:fig.2}, and the detailed explanations of the prediction model are given below.

\subsubsection{Map and Agent Encoders} 
The dynamic state of the target agent and its surrounding agents $s_{i}^{t}$ at timestep $t$ is in the format of $(x, y, v_x, v_y, \phi)$, where $(x, y)$ is the coordinate, $(v_x, v_y)$ the velocity, and $\phi$ the heading angle. Note that the coordinate system is centered on the target agent's position at the current timestep with its heading aligned with the x-axis. Therefore, the historical state of an agent $S_{i}$ can be represented by a tensor with shape $(T_h, 5)$. The agent encoder consists of two layers, i.e., one 1D convolutional layer and one long short-term memory (LSTM) layer to extract the temporal motion feature of the agent. We only consider up to ten surrounding agents within a radius of 30 meters to the target agent. All the agents including the target and surrounding agents share the same agent encoder.

The map information $\mathcal{M}$ is represented by a set of waypoints from different segments of the map. Each waypoint has a unique feature of $(x, y, \psi)$, where $(x, y)$ is the coordinate relative to the target agent, and $\psi$ is the direction. The waypoints from the same lane have the same lane features, which are the turning direction, whether the lane is in an intersection, and whether the lane has traffic control measures. These lane features are first one-hot encoded respectively and then concatenated with the waypoint features. All the waypoints share the same map encoder, which is shown in Fig. \ref{fig:fig.2}. First of all, the waypoints features are feed into a fully connected layer and we use the max-pooling operation to aggregate the information from all the waypoints on the same lane; the lane features are processed by another fully connected layer. The waypoint feature, aggregated feature, and lane feature are concatenated and passed through a fully connected layer to get the final feature vector of the waypoint. 

\subsubsection{Agent-agent Encoder}
We first represent the relationship between the agents as a graph, shown in Fig. \ref{fig:fig.1}. All the agents (nodes) are connected to the target agent (including the self-loop), and we ignore the edge attributes. We use a Transformer layer with the multi-head attention mechanism to encode the interactions between the agents, as seen in Fig. \ref{fig:fig.2}. In addition to the multi-head attention, the encoder also contains two position-wise fully connected feed-forward layers and two layer normalization layers following the Transformer architecture. The multi-head attention mechanism is illustrated as follows \cite{vaswani2017attention}.
\begin{equation}
\resizebox{0.90\linewidth}{!}{
$\mathrm{MultiHead}(\mathbf{Q}, \mathbf{K}, \mathbf{V}) = \mathrm{Concat} \left( \mathrm{head}_1, \cdots, \mathrm{head}_h \right)W^O,$
}
\end{equation}
\begin{equation}
\mathrm{head}_i = \mathrm{Attention}(\mathbf{Q}W^{Q}_i, \mathbf{K}W^{K}_i, \mathbf{V}W^{V}_i),
\end{equation}
where $h$ is the total number of attention heads, $\mathbf{Q}, \mathbf{K}, \mathbf{V}$ are the query, key, and value vectors, respectively, and $W^{Q}_i, W^{K}_i, W^{V}_i, W^{O}$ are the matrices for linear projection. The attention operation is called scaled dot-product attention, which is shown as
\begin{equation}
\mathrm{Attention}(\mathbf{Q}, \mathbf{K}, \mathbf{V}) = \mathrm{softmax}\left( \frac{\mathbf{Q}\mathbf{K}^T}{\sqrt{d_k}} \right)\mathbf{V},
\end{equation}
where $d_k$ is the dimension of the key vector.

In the agent-agent encoder, the query is the target agent's feature vector from the agent encoder and the key and value are the feature vectors of all the agents. 

\subsubsection{Agent-map Encoder} 
The relationship between the target agent and the map is represented as a graph, where the target agent can attend to all the elements in the map. We use another Transformer layer to model the agent-map relationship, as seen in Fig. \ref{fig:fig.2}, and we modify the multi-head attention to multi-modal attention as shown in Eq. \ref{multimodal}. Specifically, we do not concatenate the results from individual heads and project the concatenated vector to a low-dimensional one, but instead, we directly output the results of individual heads, and the final trajectory outputs are conditioned on the individual heads.
\begin{equation}
\label{multimodal}
\mathrm{MultiModal}(\mathbf{Q}, \mathbf{K}, \mathbf{V}) = \left( \mathrm{head}_1, \cdots, \mathrm{head}_h \right).
\end{equation}

The output of the agent-map encoder is a mode-wise feature, which means each mode has a different feature, corresponding to a different relationship between the target agent and map. To ensure the diversity of these modes, i.e., attending to different parts on the map, we only back-propagate the loss through the individual head that is closest to the ground truth in terms of final displacement error. In the agent-map encoder, the query is the interaction feature from the agent-agent encoder and the key and value are the feature vectors from the map encoder. 

\subsubsection{Score and Trajectory Decoders}
The predicted trajectories and their scores are conditioned on three features, i.e., the target agent's historical state, the interaction among agents, and the target agent's attention on the map. The interaction feature and target agent feature are first repeated along the mode axis to match with the shape of the multi-modal agent-map feature, and then the three features are concatenated to form a final representation of the driving environment. The trajectory decoder is a mode-wise four-layer MLP with the final layer outputting the coordinates of the trajectory at each timestep. The score decoder follows the same structure, except for the last layer that outputs the score of the predicted trajectories. The scores of all the predicted trajectories are grouped and passed through a softmax layer to yield a probability distribution.

\subsection{Training Objectives} 
All the modules in the prediction model are differentiable, and thus we can train the model end-to-end. To predict the trajectories, we use the smooth L1 loss on all predicted time steps. For a data point, the trajectory regression loss is defined as:
\begin{equation}
\mathcal{L}_{traj} = \sum_t \mathcal{L}_{1} \left(\mathbf{p}_{gt}^{t}- \mathbf{p}_{j^*}^{t} \right),
\end{equation}
where $\mathbf{p}_{gt}^{t}$ is the ground truth position at time step $t$. Using the variety or Minimum over N (MoN) loss \cite{thiede2019analyzing}, we only calculate the loss between the ground truth and the closest output prediction, and $j^*$ is the index of the predicted trajectory that is closest to the ground truth in terms of L2 distance between the endpoint $\mathbf{p}_{j}^{T_f}$ and the ground truth endpoint $\mathbf{p}_{gt}^{T_f}$:
\begin{equation}
j^* = \underset{j \in \{1, \dots, K\}} {\mathrm{argmin}} \parallel \mathbf{p}_{j}^{T_f} - \mathbf{p}_{gt}^{T_f} \parallel_2.
\end{equation}

Since different predicted trajectories are conditioned on the different attention heads in the agent-map encoder, only the head that corresponds to the closet trajectory gets updated, making the heads attend to different parts of the map and ensuring the diversity of the heads.

The scoring loss is the cross entropy loss between the ground truth scores (probability distribution) and the predicted probability distribution $p$. For a data point, the scoring loss is defined as:
\begin{equation}
\mathcal{L}_{score} = \mathcal{L}_{CE}(p_{gt}, p),
\end{equation}
where the ground truth distribution $p_{gt}$ is defined as:
\begin{equation}
p_{gt} = \frac{\exp - \parallel \mathbf{p}^{T_f} - \mathbf{p}_{gt}^{T_f} \parallel_2}{\sum_j \exp -\parallel \mathbf{p}_{j}^{T_f} - \mathbf{p}_{gt}^{T_f} \parallel_2}.    
\end{equation}

The total loss is a weighted sum of the trajectory regression loss and the scoring loss:
\begin{equation}
\label{loss}
\mathcal{L} = \mathcal{L}_{score} + \alpha \mathcal{L}_{traj},
\end{equation}
where $\alpha$ is a hyperparameter to balance the two learning objectives.

\section{Experiments}
\subsection{Experimental Setup}
\subsubsection{Dataset}
The proposed method is validated on the Argoverse Motion Forecasting dataset \cite{chang2019argoverse}, which contains 324,557 real-world driving scenarios for training and validation. For each scenario, five-second trajectory sequences of each tracked object sampled at 10 Hz are provided and the map information is represented as a set of lane centerlines, which is composed of a set of waypoints. The prediction task is to forecast the future possible trajectories of the target agent in a scenario over the next 3 seconds, given the 2-second historical trajectory of the target agent, and the trajectories of its neighboring agents, as well as the map context. The whole dataset is split into 205,942 training, 39,472 validation, and 78,143 testing sequences, respectively. We train the prediction model on the training set and test it on the standard testing set to benchmark the performance of the model.

\subsubsection{Metrics}
The performance of the prediction model is evaluated using some standard evaluation metrics, which are minimum average displacement error (minADE), minimum final displacement error (minFDE), brier-minFDE, and miss rate (MR). minADE and minFDE are two common distance-based metrics; minADE reports the average displacement error between the best-predicted trajectory and the ground truth over the entire time steps, and minFDE reports the displacement error at the endpoint. The best-predicted trajectory refers to the trajectory that has the minimum endpoint error. To evaluate the scoring function of the model, we employ anthor metric; brier-minFDE is defined as the sum of minFDE and the brier score $(1 - p)^2$, where $p$ is the probability of the best-predicted trajectory. Additionally, the miss rate (MR) is reported, which is defined as the ratio of the scenarios in which none of the endpoints of predicted trajectories are within 2.0 meters of ground truth.

\begin{figure*}[htp]
    \centering
    \includegraphics[width=0.98\linewidth]{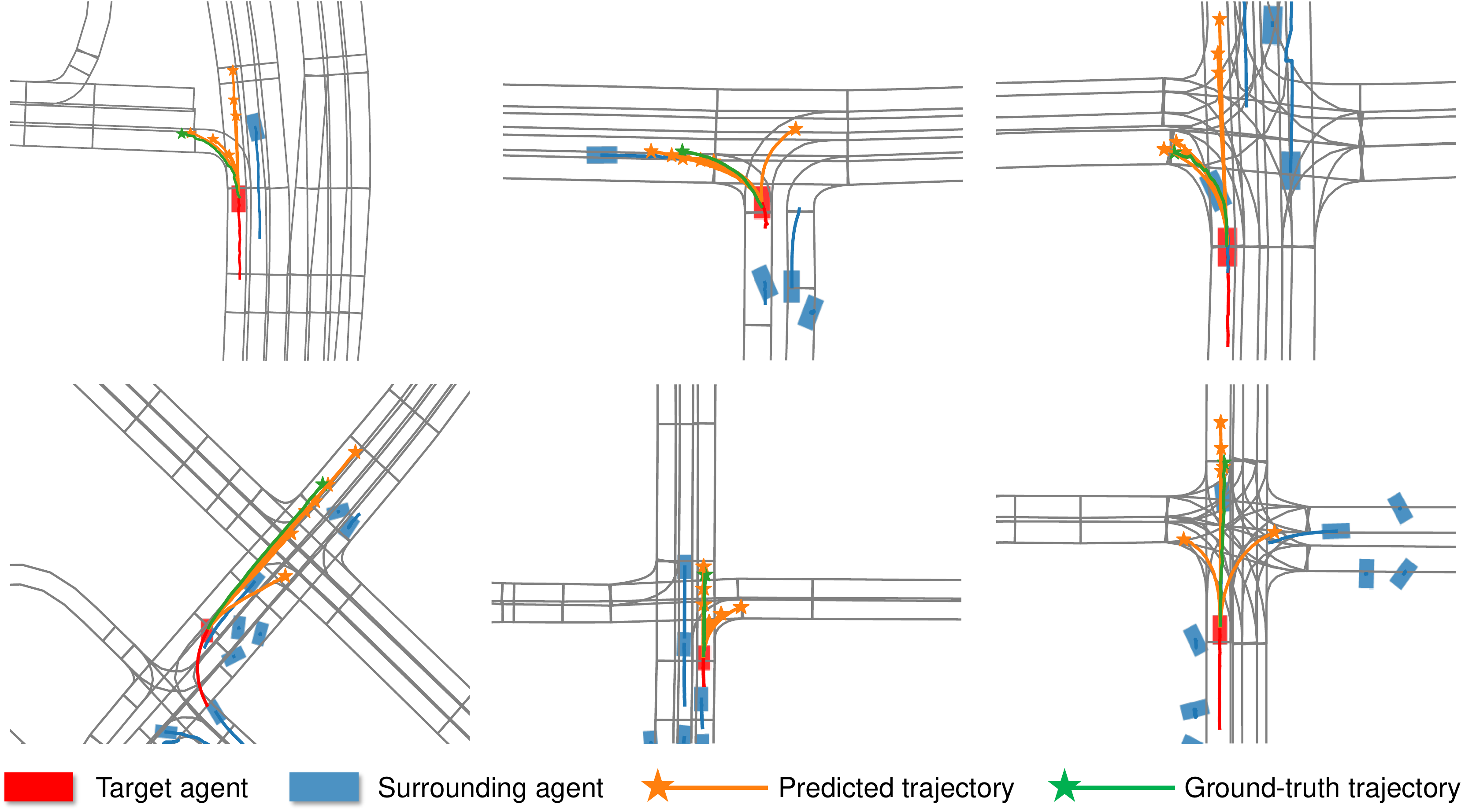}
    \caption{The qualitative motion forecasting results of the proposed model on the Argoverse validation set. The historical trajectory of the target agent is in red, and the surrounding agents are in blue; the predicted trajectories in yellow and ground truth trajectory in green, respectively.}
    \label{fig:fig.3}
\end{figure*}

\subsection{Implementation Details}
\subsubsection{Input and Output} 
For the map information, we search for 40 lanes closest to the current position of the target agent, each lane with 10 waypoints. For the agent information, we search for 10 neighboring agents within 30 meters to the target agent and organize them into a tensor along with the target agent. The missing lanes or agents in the tensors are padded with zeros and masked out when calculating the attention. The historical horizon is 2 seconds ($T_h = 20$) and the prediction horizon is 3 seconds ($T_f=30$). The output of the model is $K$ possible trajectories (a sequence of $(x,y)$ coordinates), with a matching probability for each trajectory. The number of prediction modes is set as $K=6$.

\subsubsection{Network Structure}
The dimensions of the embedded agent and map waypoint features are both 256. In the agent-agent encoder, the number of heads in the multi-head attention is 6, and the feed-forward network first projects the feature vector to 1024-dimension and then reduces it to 256-dimension. In the agent-agent encoder, the number of modes (heads) in the multi-modal attention is 6 and the remaining structure is the same as the agent-agent encoder. The final feature vector obtained from the environment encoders is with the shape of $(6, 768)$, and then feed into the trajectory and score decoders to produce the trajectory and score per mode, respectively. All the activation functions in the dense layers are ELU, and all the fully connected layers are followed by dropout layers with a dropout rate of 0.1 to mitigate overfitting. The total number of parameters of the model is 6,328,125.

\subsubsection{Training}
The hyperparameter $\alpha$ in the loss function (Eq. \ref{loss}) is 0.5 after some trials. We use Nadam optimizer with a learning rate that starts with 1e-4 and decays by 50\% after every 20 epochs. To stabilize training, we use gradient clipping with a threshold of 5 (by norm). The number of training epochs is 100 and the batch size is 64. No data augmentation is used and training one epoch on the training set takes about 10 minutes using Tensorflow with an NVIDIA RTX 2080Ti GPU. 

\subsection{Results}
\subsubsection{Qualitative Results}

Fig. \ref{fig:fig.3} shows some representative examples of the motion forecasting results given by our prediction model. The model is capable of outputting multiple possible trajectories that are diverse and compliant with the structure of the map. The best-predicted trajectory is very close to the ground-truth one while the model maintains the ability to predict other possible trajectories with varying speed profiles or directions. The qualitative results demonstrate the effectiveness of our proposed model on different complex urban driving scenarios including left-turn, right-turn, intersection, etc. 

\begin{table*}[htp]
\centering
\caption{The quantitative results in comparison with existing methods on the Argoverse benchmark (test set)}
%\resizebox{0.7\linewidth}{!}{%
\begin{tabular}{ccccc}
\hline
Method                                      & minADE (m)      & minFDE(m)       & brier-minFDE  & miss rate       \\ \hline
PRIME \cite{song2021learning}               & 1.2187          & 1.5582          & 2.0978        & 0.1150         \\
LaneRCNN \cite{zeng2021lanercnn}            & 0.9038          & 1.4526          & 2.1470        & 0.1232      \\
TNT \cite{zhao2020tnt}                      & 0.9097          & 1.4457          & 2.1401        & 0.1656         \\
Multi-head attention \cite{mercat2020multi} & 0.9973          & 1.4209          & 2.1154        & 0.1308         \\
LaneGCN \cite{liang2020learning}            & 0.8679          & 1.3550          & 2.0495        & 0.1597                \\
mmTransformer \cite{liu2021multimodal}      & 0.8436          & 1.3383          & 2.0328        & 0.1540                 \\
HOME \cite{gilles2021home}                  & 0.8904          & 1.2919          &\textbf{1.8601}& \textbf{0.0846}   \\
Ours (Multi-modal Transformer)              & \textbf{0.8372} & \textbf{1.2905} & 1.9393        & 0.1429 \\ \hline
\end{tabular}%
%}
\label{tab1}
\end{table*}

\begin{figure}[htp]
    \centering
    \includegraphics[width=0.96\linewidth]{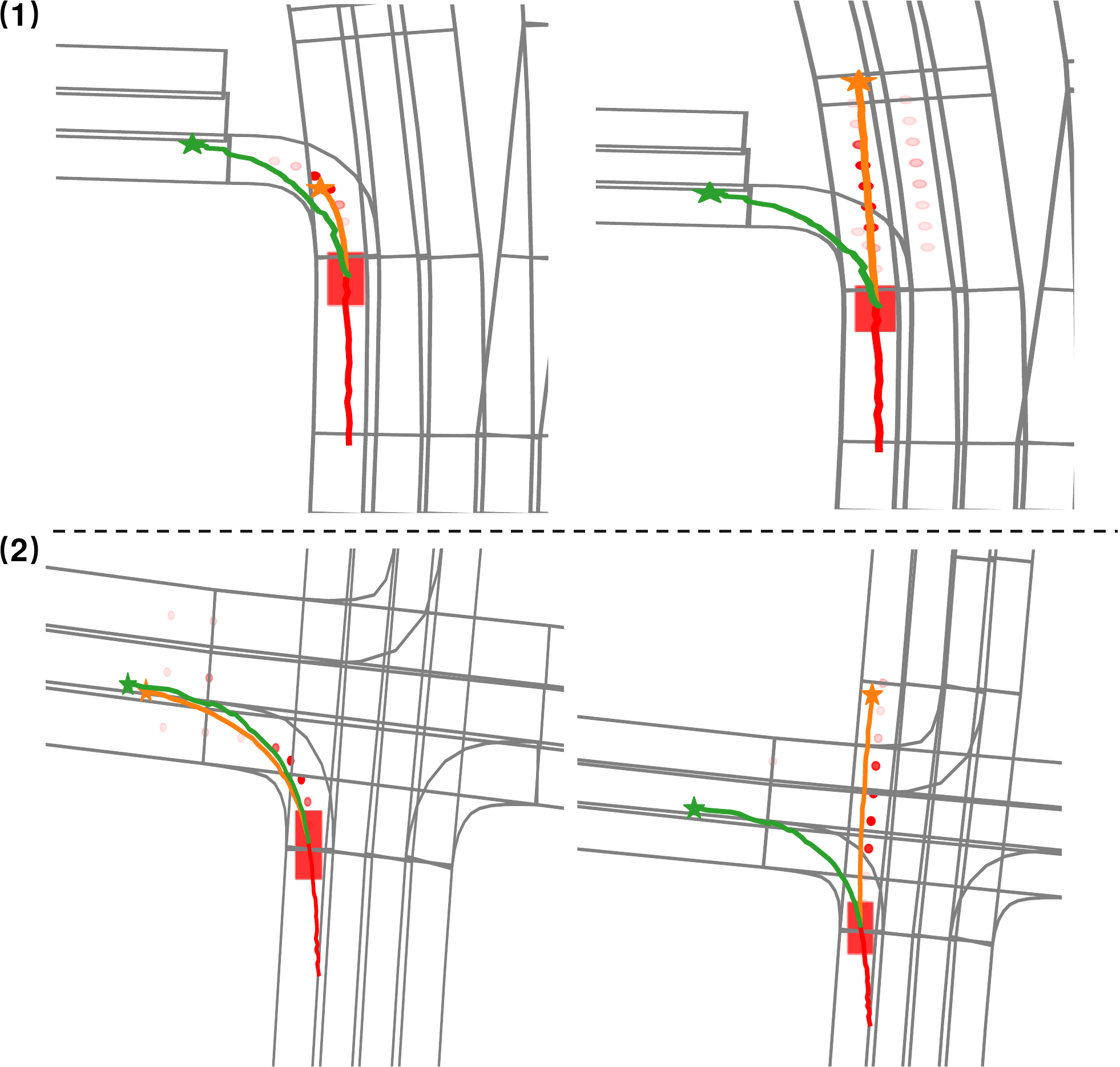}
    \caption{The visualization of the attention scores on the map waypoints for different modes. The darker red means more attention on the waypoint.}
    \label{fig:fig.4}
\end{figure}

Here, we visualize the target agent's attention on the map waypoints, as seen in Fig. \ref{fig:fig.4}, to demonstrate the interpretability of our prediction model. The attention of each mode (head) to the map waypoints is represented as attention scores, and the waypoints with scores greater than 0.01 are displayed on the map according to different modes. In the given two cases of left-turn scenarios, we can notice that the more attention on the left-turn lane when the model predicts a left-turn trajectory, and likewise, more attention on the straight lane when the model predicts a go-straight trajectory. The results manifest that the proposed multi-modal attention can automatically learn to extract the possible goals on the map, and thus the predicted trajectories can be conditioned on the different goals. 

\subsubsection{Quantitative Results}

The proposed method is evaluated with the quantitative metrics previously defined and compared against the state-of-the-art methods on the Argoverse benchmark (test set). We only report the results with six forecasted trajectories ($K=6$), which are summarized in Table \ref{tab1}. Our proposed method can achieve the best prediction accuracy in terms of the two distance error-based metrics (minADE and minFDE). The brier-minFDE and miss rate of our model are slightly worse than that of HOME \cite{gilles2021home}. This is because HOME is optimized for minimizing the miss rate, and the weight of the scoring term in the loss function needs further tuning in training our model to improve the scoring performance. It is also worth noting that our model is with a smaller size and simpler training process, which can ease the burden of pre- or post-processing and simplify the model training process, as well as bring fast inference.

\subsection{Ablation Study}
We conduct an ablation study to evaluate and analyze the influence of map structure and the contributions of our proposed multi-modal attention to the final prediction accuracy. For the map structure, we investigate two different levels of representations, i.e., lane and waypoint. To encode the feature of a map lane, we use global max-pooling to aggregate the features of all the waypoints in a lane. For multi-modal prediction, in addition to our proposed multi-modal attention, we take the ensemble method trained with the variety loss as a comparison. It uses an ensemble of trajectory decoders to output different trajectories from the same extracted environment feature vector. All these methods are tested on the Argoverse standard test set with six predicted trajectories. From the results given in Table \ref{tab2}, we can conclude that our proposed method with map waypoints and multi-modal agent-map attention can delivery the best prediction accuracy. Using the map waypoints can bring a higher map resolution and reduce the loss of information. Moreover, using the proposed multi-modal attention can achieve not only better prediction accuracy but also better interpretability, as shown in the previous section.

\begin{table}[htp]
\centering
\caption{Ablation study of the map structure and multi-modal prediction method on the Argoverse test set}
\resizebox{0.95\linewidth}{!}{%
\begin{tabular}{@{}cc|cc|cc@{}}
\toprule
\multicolumn{2}{c|}{Map}                              & \multicolumn{2}{c|}{Multi-modal}                      & \multirow{2}{*}{minADE (m)} & \multirow{2}{*}{minFDE(m)} \\ \cmidrule(r){1-4}
Lane                      & Waypoint                  & Attention                 & Ensemble                  &                             &                            \\ \midrule
& \checkmark              & \checkmark                &                                                       & \textbf{0.8372}             & \textbf{1.2905}            \\ \midrule
\checkmark                &                           & \checkmark                &                           & 0.8461                      & 1.3097                    \\ \midrule
                          & \checkmark                &                           & \checkmark                & 0.8512                      & 1.3199                   \\ \midrule
\checkmark                &                           &                           & \checkmark                & 0.8604                      & 1.3373                     \\ \bottomrule
\end{tabular}%
}
\label{tab2}
\end{table}

\section{CONCLUSIONS}

In this paper, we propose a multi-modal trajectory prediction model based on the Transformer structure. We employ a multi-head attention Transformer layer to model the relationship among interacting agents and introduce a multi-modal attention Transformer layer to extract the different relationships between the target agent and map waypoints, which determines the final trajectory outputs. Comprehensive experiments on the Argoverse motion dataset reveal the effectiveness of our model with competitive accuracy, better interpretability, yet a simple structure and training process.

%%%%%%%%%%%%%%%%%%%%%%%%%%%%%%%%%%%%%%%%%%%%%%%%%%%%%%%%%%%%%%%%%%%%%%%%%%%%%%%
\bibliographystyle{IEEEtran}
\bibliography{IEEEexample}

\begin{thebibliography}{10}
\providecommand{\url}[1]{#1}
\csname url@rmstyle\endcsname
\providecommand{\newblock}{\relax}
\providecommand{\bibinfo}[2]{#2}
\providecommand\BIBentrySTDinterwordspacing{\spaceskip=0pt\relax}
\providecommand\BIBentryALTinterwordstretchfactor{4}
\providecommand\BIBentryALTinterwordspacing{\spaceskip=\fontdimen2\font plus
\BIBentryALTinterwordstretchfactor\fontdimen3\font minus
  \fontdimen4\font\relax}
\providecommand\BIBforeignlanguage[2]{{%
\expandafter\ifx\csname l@#1\endcsname\relax
\typeout{** WARNING: IEEEtran.bst: No hyphenation pattern has been}%
\typeout{** loaded for the language `#1'. Using the pattern for}%
\typeout{** the default language instead.}%
\else
\language=\csname l@#1\endcsname
\fi
#2}}

\bibitem{huang2021driving}
Z.~Huang, J.~Wu, and C.~Lv, ``Driving behavior modeling using naturalistic
  human driving data with inverse reinforcement learning,'' \emph{IEEE
  Transactions on Intelligent Transportation Systems}, 2021.

\bibitem{mozaffari2020deep}
S.~Mozaffari, O.~Y. Al-Jarrah, M.~Dianati, P.~Jennings, and A.~Mouzakitis,
  ``Deep learning-based vehicle behavior prediction for autonomous driving
  applications: A review,'' \emph{IEEE Transactions on Intelligent
  Transportation Systems}, 2020.

\bibitem{gao2020vectornet}
J.~Gao, C.~Sun, H.~Zhao, Y.~Shen, D.~Anguelov, C.~Li, and C.~Schmid,
  ``Vectornet: Encoding hd maps and agent dynamics from vectorized
  representation,'' in \emph{Proceedings of the IEEE/CVF Conference on Computer
  Vision and Pattern Recognition}, 2020, pp. 11\,525--11\,533.

\bibitem{zhao2020tnt}
H.~Zhao, J.~Gao, T.~Lan, C.~Sun, B.~Sapp, B.~Varadarajan, Y.~Shen, Y.~Shen,
  Y.~Chai, C.~Schmid, \emph{et~al.}, ``Tnt: Target-driven trajectory
  prediction,'' in \emph{Conference on Robot Learning (CoRL)}, 2020.

\bibitem{cui2019multimodal}
H.~Cui, V.~Radosavljevic, F.-C. Chou, T.-H. Lin, T.~Nguyen, T.-K. Huang,
  J.~Schneider, and N.~Djuric, ``Multimodal trajectory predictions for
  autonomous driving using deep convolutional networks,'' in \emph{2019
  International Conference on Robotics and Automation (ICRA)}.\hskip 1em plus
  0.5em minus 0.4em\relax IEEE, 2019, pp. 2090--2096.

\bibitem{salzmann2020trajectron++}
T.~Salzmann, B.~Ivanovic, P.~Chakravarty, and M.~Pavone, ``Trajectron++:
  Dynamically-feasible trajectory forecasting with heterogeneous data,'' in
  \emph{Computer Vision--ECCV 2020: 16th European Conference, Glasgow, UK,
  August 23--28, 2020, Proceedings, Part XVIII 16}.\hskip 1em plus 0.5em minus
  0.4em\relax Springer, 2020, pp. 683--700.

\bibitem{dong2021multi}
B.~Dong, H.~Liu, Y.~Bai, J.~Lin, Z.~Xu, X.~Xu, and Q.~Kong, ``Multi-modal
  trajectory prediction for autonomous driving with semantic map and dynamic
  graph attention network,'' in \emph{Machine Learning for Autonomous Driving
  Workshop at the 34th Conference on Neural Information Processing Systems},
  2021.

\bibitem{liang2020learning}
M.~Liang, B.~Yang, R.~Hu, Y.~Chen, R.~Liao, S.~Feng, and R.~Urtasun, ``Learning
  lane graph representations for motion forecasting,'' in \emph{European
  Conference on Computer Vision}, 2020, pp. 541--556.

\bibitem{lee2017desire}
N.~Lee, W.~Choi, P.~Vernaza, C.~B. Choy, P.~H. Torr, and M.~Chandraker,
  ``Desire: Distant future prediction in dynamic scenes with interacting
  agents,'' in \emph{Proceedings of the IEEE Conference on Computer Vision and
  Pattern Recognition}, 2017, pp. 336--345.

\bibitem{ivanovic2020multimodal}
B.~Ivanovic, K.~Leung, E.~Schmerling, and M.~Pavone, ``Multimodal deep
  generative models for trajectory prediction: A conditional variational
  autoencoder approach,'' \emph{IEEE Robotics and Automation Letters}, vol.~6,
  no.~2, pp. 295--302, 2020.

\bibitem{gupta2018social}
A.~Gupta, J.~Johnson, L.~Fei-Fei, S.~Savarese, and A.~Alahi, ``Social gan:
  Socially acceptable trajectories with generative adversarial networks,'' in
  \emph{Proceedings of the IEEE Conference on Computer Vision and Pattern
  Recognition}, 2018, pp. 2255--2264.

\bibitem{ye2021tpcn}
M.~Ye, T.~Cao, and Q.~Chen, ``Tpcn: Temporal point cloud networks for motion
  forecasting,'' in \emph{Proceedings of the IEEE/CVF Conference on Computer
  Vision and Pattern Recognition}, 2021, pp. 11\,318--11\,327.

\bibitem{thiede2019analyzing}
L.~A. Thiede and P.~P. Brahma, ``Analyzing the variety loss in the context of
  probabilistic trajectory prediction,'' in \emph{Proceedings of the IEEE/CVF
  International Conference on Computer Vision}, 2019, pp. 9954--9963.

\bibitem{chai2020multipath}
Y.~Chai, B.~Sapp, M.~Bansal, and D.~Anguelov, ``Multipath: Multiple
  probabilistic anchor trajectory hypotheses for behavior prediction,'' in
  \emph{Conference on Robot Learning}, 2020.

\bibitem{phan2020covernet}
T.~Phan-Minh, E.~C. Grigore, F.~A. Boulton, O.~Beijbom, and E.~M. Wolff,
  ``Covernet: Multimodal behavior prediction using trajectory sets,'' in
  \emph{Proceedings of the IEEE/CVF Conference on Computer Vision and Pattern
  Recognition}, 2020, pp. 14\,074--14\,083.

\bibitem{song2021learning}
H.~Song, D.~Luan, W.~Ding, M.~Y. Wang, and Q.~Chen, ``Learning to predict
  vehicle trajectories with model-based planning,'' \emph{arXiv preprint
  arXiv:2103.04027}, 2021.

\bibitem{zhang2020map}
L.~Zhang, P.-H. Su, J.~Hoang, G.~C. Haynes, and M.~Marchetti-Bowick,
  ``Map-adaptive goal-based trajectory prediction,'' in \emph{Conference on
  Robot Learning}, 2020.

\bibitem{liu2021multimodal}
Y.~Liu, J.~Zhang, L.~Fang, Q.~Jiang, and B.~Zhou, ``Multimodal motion
  prediction with stacked transformers,'' in \emph{Proceedings of the IEEE/CVF
  Conference on Computer Vision and Pattern Recognition}, 2021, pp. 7577--7586.

\bibitem{fang2020tpnet}
L.~Fang, Q.~Jiang, J.~Shi, and B.~Zhou, ``Tpnet: Trajectory proposal network
  for motion prediction,'' in \emph{Proceedings of the IEEE/CVF Conference on
  Computer Vision and Pattern Recognition}, 2020, pp. 6797--6806.

\bibitem{vaswani2017attention}
A.~Vaswani, N.~Shazeer, N.~Parmar, J.~Uszkoreit, L.~Jones, A.~N. Gomez,
  {\L}.~Kaiser, and I.~Polosukhin, ``Attention is all you need,'' in
  \emph{Advances in neural information processing systems}, 2017, pp.
  5998--6008.

\bibitem{chang2019argoverse}
M.-F. Chang, J.~Lambert, P.~Sangkloy, J.~Singh, S.~Bak, A.~Hartnett, D.~Wang,
  P.~Carr, S.~Lucey, D.~Ramanan, \emph{et~al.}, ``Argoverse: 3d tracking and
  forecasting with rich maps,'' in \emph{Proceedings of the IEEE/CVF Conference
  on Computer Vision and Pattern Recognition}, 2019, pp. 8748--8757.

\bibitem{zeng2021lanercnn}
W.~Zeng, M.~Liang, R.~Liao, and R.~Urtasun, ``Lanercnn: Distributed
  representations for graph-centric motion forecasting,'' \emph{arXiv preprint
  arXiv:2101.06653}, 2021.

\bibitem{mercat2020multi}
J.~Mercat, T.~Gilles, N.~El~Zoghby, G.~Sandou, D.~Beauvois, and G.~P. Gil,
  ``Multi-head attention for multi-modal joint vehicle motion forecasting,'' in
  \emph{2020 IEEE International Conference on Robotics and Automation
  (ICRA)}.\hskip 1em plus 0.5em minus 0.4em\relax IEEE, 2020, pp. 9638--9644.

\bibitem{gilles2021home}
T.~Gilles, S.~Sabatini, D.~Tsishkou, B.~Stanciulescu, and F.~Moutarde, ``Home:
  Heatmap output for future motion estimation,'' \emph{arXiv preprint
  arXiv:2105.10968}, 2021.

\end{thebibliography}
\end{document}